\documentclass{sig-alternate-2013}
\usepackage[outdir=./]{epstopdf}
\usepackage{multirow}

\newfont{\mycrnotice}{ptmr8t at 7pt}
\newfont{\myconfname}{ptmri8t at 7pt}
\let\crnotice\mycrnotice%
\let\confname\myconfname%

\permission{Permission to make digital or hard copies of all or part of this work for personal or classroom use is granted without fee provided that copies are not made or distributed for profit or commercial advantage and that copies bear this notice and the full citation on the first page. Copyrights for components of this work owned by others than ACM must be honored. Abstracting with credit is permitted. To copy otherwise, or republish, to post on servers or to redistribute to lists, requires prior specific permission and/or a fee. Request permissions from Permissions@acm.org.}
\conferenceinfo{MM'15,}{October 26--30, 2015, Brisbane, Australia.}
\copyrightetc{\copyright~2015 ACM. ISBN \the\acmcopyr}
\crdata{978-1-4503-3459-4/15/10\ ...\$15.00.\\
DOI: http://dx.doi.org/10.1145/2733373.2809930}

\begin{document}
%

\title{Who are the Devils Wearing Prada in New York City?}
%
%
%
%
%

\numberofauthors{5} 
%
\author{
%
%
\alignauthor
KuanTing Chen\\
       \affaddr{National Taiwan University, Taipei, Taiwan}\\
       \affaddr{ktchen@cmlab.csie.ntu.edu.tw}
\alignauthor
Kezhen Chen\thanks{KuanTing Chen and Kezhen Chen contributed equally to this work.}\\
       \affaddr{University of Rochester, Rochester, New York, USA}\\
       \affaddr{kchen33@u.rochester.edu}
\alignauthor 
Peizhong Cong\\
       \affaddr{University of Rochester, Rochester, New York, USA}\\
       \affaddr{cong314159@gmail.com}
\and  
\alignauthor Winston H. Hsu\\
       \affaddr{National Taiwan University, Taipei, Taiwan}\\
       \affaddr{whsu@ntu.edu.tw}
\alignauthor Jiebo Luo\\
       \affaddr{University of Rochester, Rochester, New York, USA}\\
       \affaddr{jluo@cs.rochester.edu}
}
 
\maketitle
\begin{abstract}
Fashion is a perpetual topic in human social life, and the mass has the penchant to emulate what large city residents and celebrities wear. Undeniably, New York City is such a bellwether large city with all kinds of fashion leadership. Consequently, to study what the fashion trends are during this year, it is very helpful to learn the fashion trends of New York City. Discovering fashion trends in New York City could boost many applications such as clothing recommendation and advertising. Does the fashion trend in the New York Fashion Show actually influence the clothing styles on the public? To answer this question, we design a novel system that consists of three major components: (1) constructing a large dataset from the New York Fashion Shows and New York street chic in order to understand the likely clothing fashion trends in New York, (2) utilizing a learning-based approach to discover fashion attributes as the representative characteristics of fashion trends, and (3) comparing the analysis results from the New York Fashion Shows and street-chic images to verify whether the fashion shows have actual influence on the people in New York City. Through the preliminary experiments over a large clothing dataset, we demonstrate the effectiveness of our proposed system, and obtain useful insights on fashion trends and fashion influence. 

\end{abstract}




\section{Introduction}
\begin{figure}
\label{fig:idea}
\centering
\epsfig{file=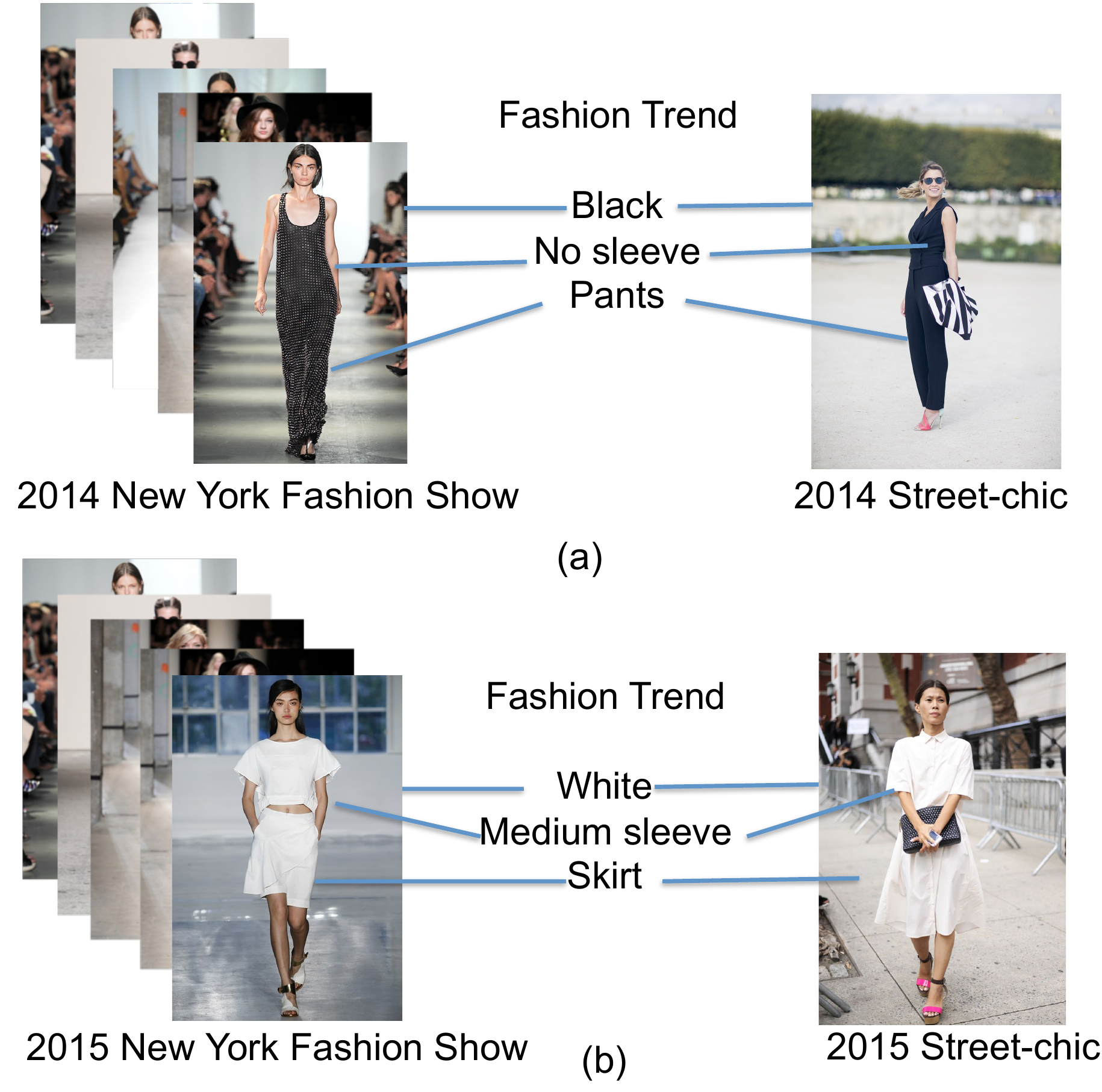, height=1.9in, width=2.2in}
\caption{Examples for the influence of New York Fashion Show on the public.}
\end{figure}

"The devil wears Prada" is a very popular fashion related movie which describes the story between a recent college graduated girl and the diabolic fashion magazine editor Miranda Priestly. Although Miranda's grim manner of management is impressive, we also admire her unique foresight and aesthetics of fashion. In modern times, a growing number of people pay more attention to fashion and the mass has the penchant to emulate what large city residents and celebrities wear. Moreover, investigating fashion trends has become a great interest for the industry and academia \cite{RW:BourdevMM11}\cite{RW:LiuCVPR12}\cite{RW:NguyenMM12}\cite{RW:TMM14} because of its promising opportunity for boosting many emerging applications such as clothing recommendation, advertising by clothing brand association, etc. Consequently, learning the fashion trends every year is very important for the industry as well as sociology and psychology. 

Indisputably, New York City is one of the most famous fashion metropolises and the New York Fashion Week is one of the four most significant fashion weeks around the world \cite{report:NYT}\cite{report:BBC}\cite{report:fashiontrend}. The New York Fashion Show consists of almost 300 catwalk shows which are attended by more than 100,000 people \cite{report:NYT}. The New York City Economic Development Corporation reported that the local economy is boosted by 850 million dollars by the New York Fashion Week \cite{report:NYCEDE}.

A traditional way to discover fashion trend in the New York Fashion Week would be relying on the domain knowledge of the clothing designers and experts. However, it is very time consuming and the fashion trends vary with the season. Hidayati et al. \cite{hidayati:TrendNY} proposed to exploit fashion elements by clustering visual patches in the video frames of the fashion show. In contrast to this work, we focus on learning the clothing fashion trends by fashion attribute discovery. Moreover, to our best knowledge, this is the first work to address the question "Does the fashion trend in the New York Fashion Show actually influence the clothing styles of the public?" (cf. Figure \ref{fig:idea}).

In this paper, we first construct a large clothing dataset from two resources: (1) a large number of images from fashion shows during the annual New York Fashion Week. (2) Street-chic images after New York Fashion Shows in order to gauge whether the fashion trends have influence on people's daily life. Next, we analyze 2014 and 2015 fashion show images and extract the fashion attributes. More specifically, for every image, we automatically extract 60 clothing attributes such as collar, necktie, pants, etc. Using semantic clothing attributes to represent fashion trends can tell people the most popular style, which is not only a good way to learn fashion trends for research, but also is a specific fashion reference for people to choose. Finally, we utilize these attributes as our predicted clothing fashion style and compare these attributes with attributes from street-chic. In our experimental results, we demonstrate the effectiveness of the  clothing attributes and the correlation of clothing style between fashion shows and street-chic.

\begin{figure}
\centering

\epsfig{file=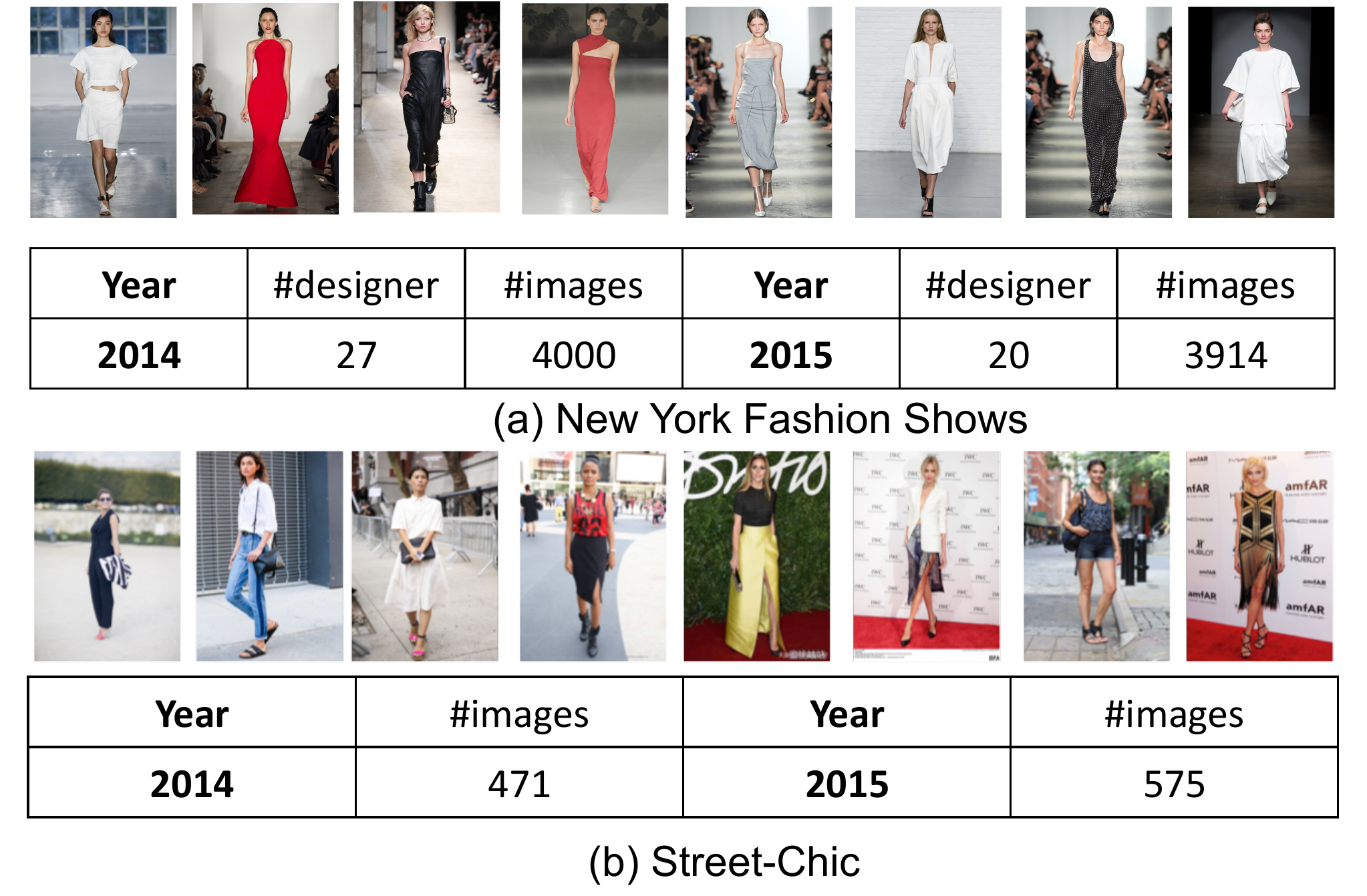, height=1.6in, width=3in}
\caption{Examples from two data resources for clothing style analysis, (a) New York Fashion Show and (b) Street-chic. \label{fig:datasets}}
\end{figure}

\section{Dataset Construction}
In this work, we mainly exploit the fashion trend in 2015 and the influence of fashion trends on people's daily life. In our dataset, in order to study the fashion influence on people's daily life, we collect two datasets. \textbf{(1) Street-chic dataset.} This dataset contains 471 street-chic images in New York from April 2014 to July 2014 and 575 street-chic images in New York from April 2015 to July 2015 \cite{dataset:Pinterest}\cite{dataset:flickr}. It is worth noting that the number of images from street chic is not very large in each year, but these images are high quality and representative (cf. Figure \ref{fig:datasets})    \textbf{(2) New York Fashion Show.} As these street-chic images are all from summer/spring season, we collected 3914 images from 2014 summer/spring New York Fashion Show and 4000 images from  2015 summer/spring New York Fashion Show, respectively \cite{dataset:vogue}. The 7914 images from 2014 and 2015 New York Fashion Shows are used to extract features and fashion attributes to analyze fashion trends. We summarize and illustrate examples of our datasets in Figure \ref{fig:datasets}.

\section{Fashion Style Discovery}

To provide a significant and specific clothing fashion reference, an effective framework for exploring clothing style representation is required. Using an appropriate clothing representation, we could offer a semantic and intuitive way for people to determine what fashion style they would like to choose. Inspired by the paper \cite{Chen:clothAttri}, a learning-based clothing attributes approach was carried out to describe clothing style. In the research \cite{Chen:clothAttri}, Chen et al. only detected 42 upper body clothing attributes. However, we observe the clothing information in the lower body is an essential clue for fashion trends understanding (e.g. wearing pants or skirt). Different from this research \cite{Chen:clothAttri}, we proposed to exploit a 60-attribute semantic representation to describe both upper and lower body clothing attributes. In the following, we briefly describe the adopted approaches for learning the clothing attribute. Table \ref{tab:attri} shows the details of clothing attributes.

\subsection{Feature Extraction}
\label{subsec:features}
\begin{figure}
\centering
\epsfig{file=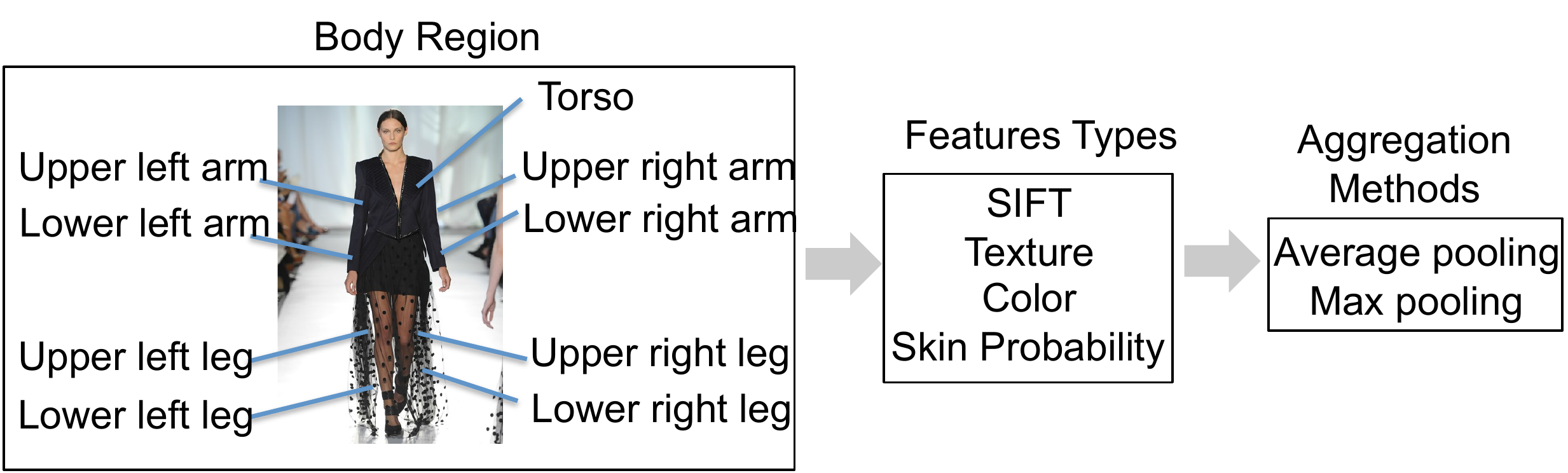, height=0.6in, width=2.2in}
\caption{An illustration of the visual features we extracted from photos. \label{fig:features}}
\end{figure}

Before learning the 60 clothing attributes, we need to extract clothing features beforehand to train classifiers for every attribute. Thanks for Marcin Eichner's team \cite{Eichner:pose}, we apply their pose estimation software to detecting the model's pose and retrieve the body region of the model. More clearly, the pose estimation process is to segment body into nine parts: torso, upper left arm, upper right arm, lower left arm, lower right arm, upper left leg, upper right leg, lower left leg and lower right leg. Next, we extract four different kinds of features (e.g. color, texture, SIFT, and skin) and utilize two different aggregation methods to generate a visual feature vector for all the parts of the body. An example is illustrated in Figure \ref{fig:features}.

\subsection{Fashion Attribute Learning}
\label{subsec:approach_Attri}
In order to determine whether an fashion element is present in an image, we adopt a Support Vector Machine (SVM) \cite{chang:svm} with a Chi-square kernel to learning fashion attributes (cf. Table \ref{tab:attri}).  The most intuitive way for training each fashion attribute is concatenating 72 features (cf. Section \ref{subsec:features}) into a long vector that becomes the full body visual feature vector. However, the influence of different types of feature on each attributes may vary. For example, texture features might have a great effect on pattern based attributes. Consequently, we compute the classification performance of each feature to represent the importance of features as weights towards individual attributes. As a result, a weighting parameter can be applied to vectors from different features to emphasize the importance of different features.

\begin{table}
\centering
\caption{A summary of Fashion attributes. \label{tab:attri}}
\begin{tabular}{c}
\raisebox{-0.4\height}{\includegraphics[width=0.44\textwidth ]{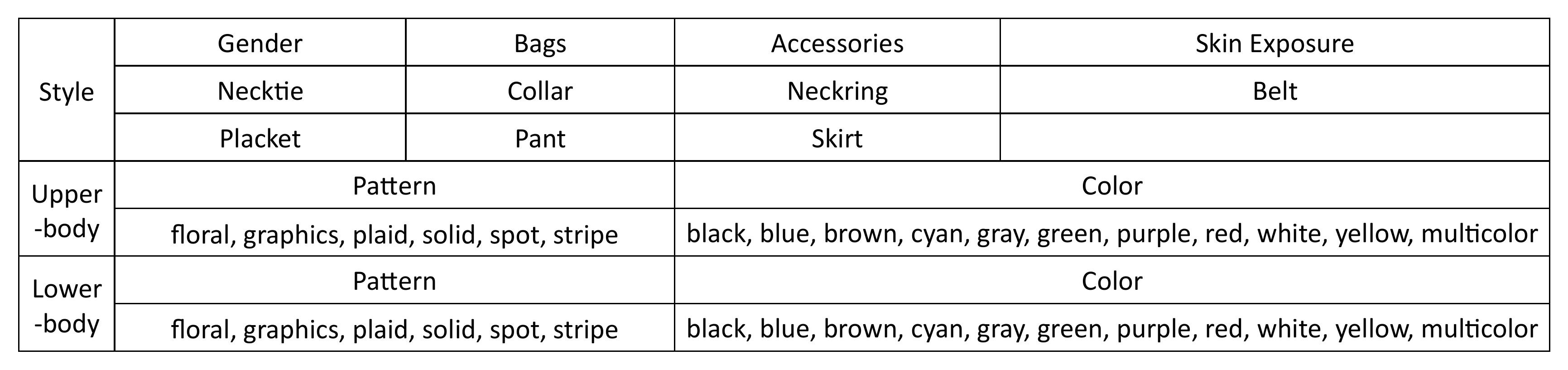}}
\end{tabular}
\end{table}

\subsection{Attribute Relation Inference}
In section \ref{subsec:approach_Attri} the fashion elements are considered as isolated attributes. However, it is highly possible that some attributes appear in pairs (or groups). For example, we observe that a plaid shirt might have more than two colors. Note that inter-attribute dependencies are not always symmetric. For example, while a plaid shirt strongly suggests the presence of more than two colors, more than two colors do not necessarily suggest a shirt being plaid. We adopt a Conditional Random Fields (CRF) approach to inference the relation betweens attributes. More specifically, each attribute acts as a node in the CRF framework and the edge connecting every two nodes indicates the joint probability of these two attributes. We build a fully connected CRF with all the attributes pairwise connected. 

\section{Experimental Results}
We conduct experiments on a large clothing dataset collected from the New York Fashion Shows and Street-chic photos. 

We conduct several experiments to gain understanding of the performance of the attributes. The overall accuracy of attributes is 62.6\%. The examples in Figure \ref{fig:idea} indicate that the extracted attributes was effective for clothing style representation. An interesting observation is that some categories suffer from worse results (e.g. 42\% accuracy of belt, 56\% accuracy of accessories) since the objects are relatively small compared to the entire body. In the future, we could segment the body into more parts to improve the accuracy of the attributes detection. For example, we could segment the middle body for belt and the neck part for the accessories.

Table \ref{tab:trend} shows the attributes percentage detected on the 2014 New York Fashion Shows and 2015 New York Fashion Shows. We divide these attributes into three categories, color, pattern and clothing style. We observe that there are always some classic colors, patterns or styles that have a large amount of clothing images in the fashion shows of every year. For example, in upper body color, both 2014 and 2015 fashion images have a large amount of white, gray and black colors. In upper body and lower body patterns, solid pattern is the classic pattern and solid pattern clothes always dominate every year's fashion shows. In the style attribute category, there are many images with skirt in spring/summer fashion shows. These examples indicate that the results generated by attribute description methods are very reasonable and these results can be a very significant reflection for the fashion shows styles.

\begin{table}
\centering
\caption{The summary of attributes percentage appeared in spring/summer fashion shows.\label{tab:trend}}
\begin{tabular}{c}
\raisebox{-0.7\height}{\includegraphics[width=0.33\textwidth]{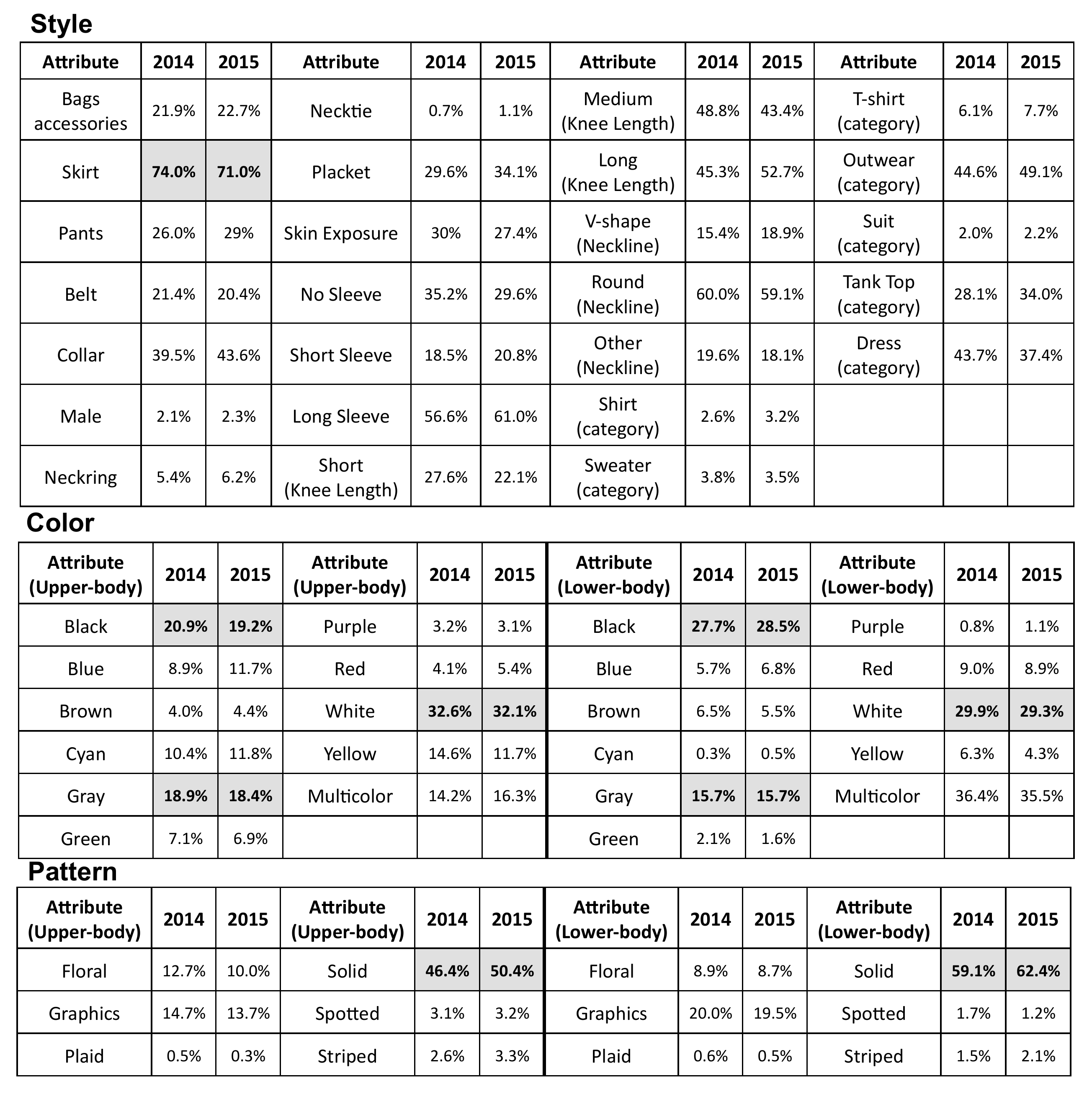}}\\
\end{tabular}
\end{table}

\begin{figure}
\centering
\epsfig{file=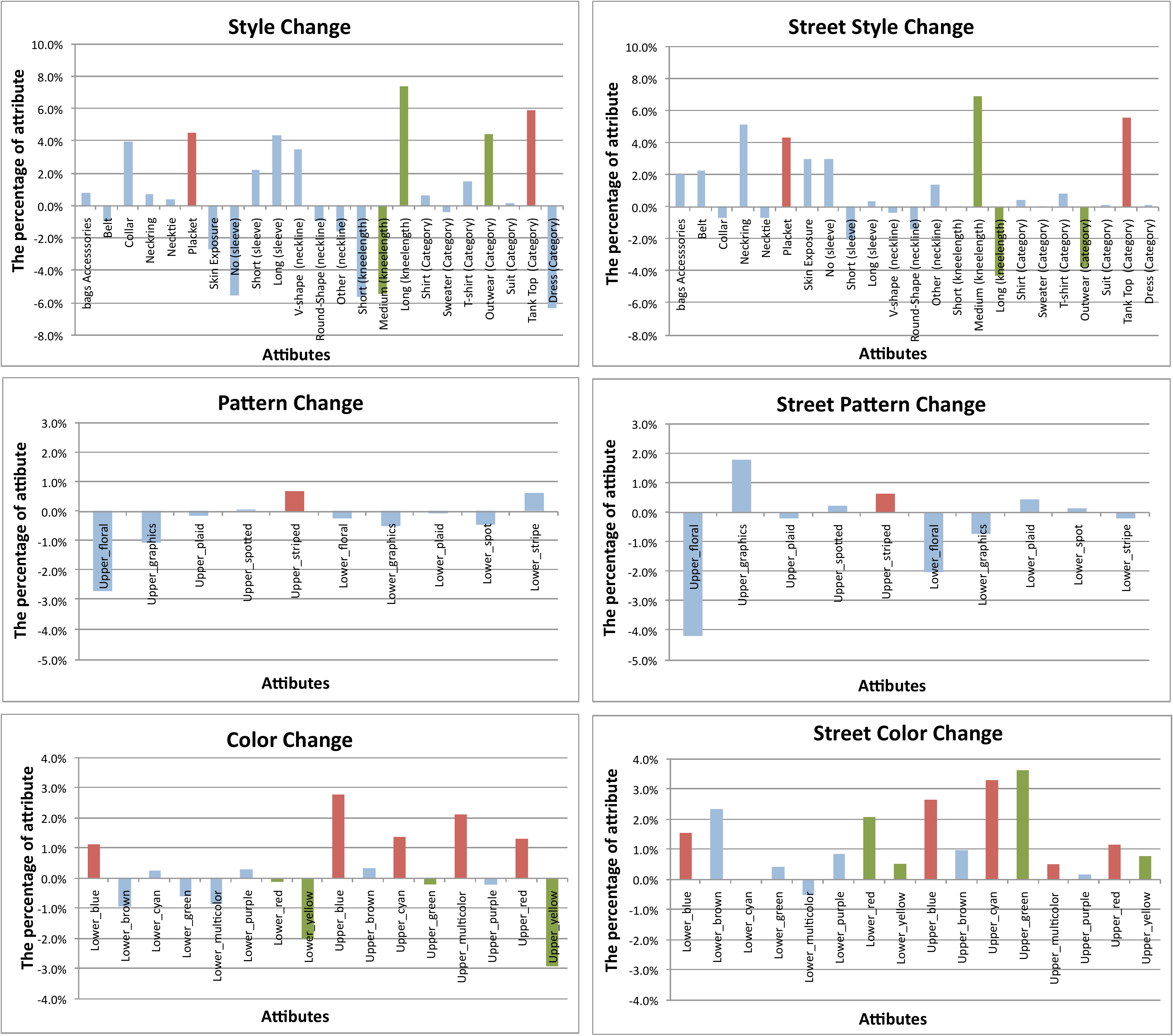, height=2.2in, width=3.2in}
\caption{Quantitative fashion style changes in 2015 shown in the New York Fashion Shows (left column) and street-chic (right column). The correlation coefficients between the two columns are (style: -0.32, pattern: 0.64, color: 0.28). Note that the positively correlated attributes are marked red and the negatively correlated attributes are marked green. In terms of style, there are positively correlated attributes even though the overall correlation is negative, revealing both correlation and distance between stage and reality.\label{fig:eval}}
\end{figure}

Besides these classic cloth styles, we are also interested in the changes in the fashion trend. These changes in attributes could indicate special fashion trends for the specific year. Note that we remove the classic style attributes, black, white and gray in the color category, solid pattern in the pattern category and skirts in the style category to present the changes more clearly. Figure \ref{fig:eval} shows the change of fashion trend between the 2015 fashion shows and the 2014 fashion shows. From this figure, we could tell the unique fashion trends for 2015. In the color category, attributes with larger changes in the upper body are blue, multicolor, cyan and red; attributes with larger changes in the lower body are blue and purple. In the pattern category, attribute with larger changes in the upper body is striped; attribute with larger changes in the lower body is striped. In the style category, long knee length, tank top, placket, outwear and collar are most significant attributes in 2015. These attributes with larger changes could be regarded as the unique fashion attributes trends shown in 2015.
 
\begin{figure}
\centering
\epsfig{file=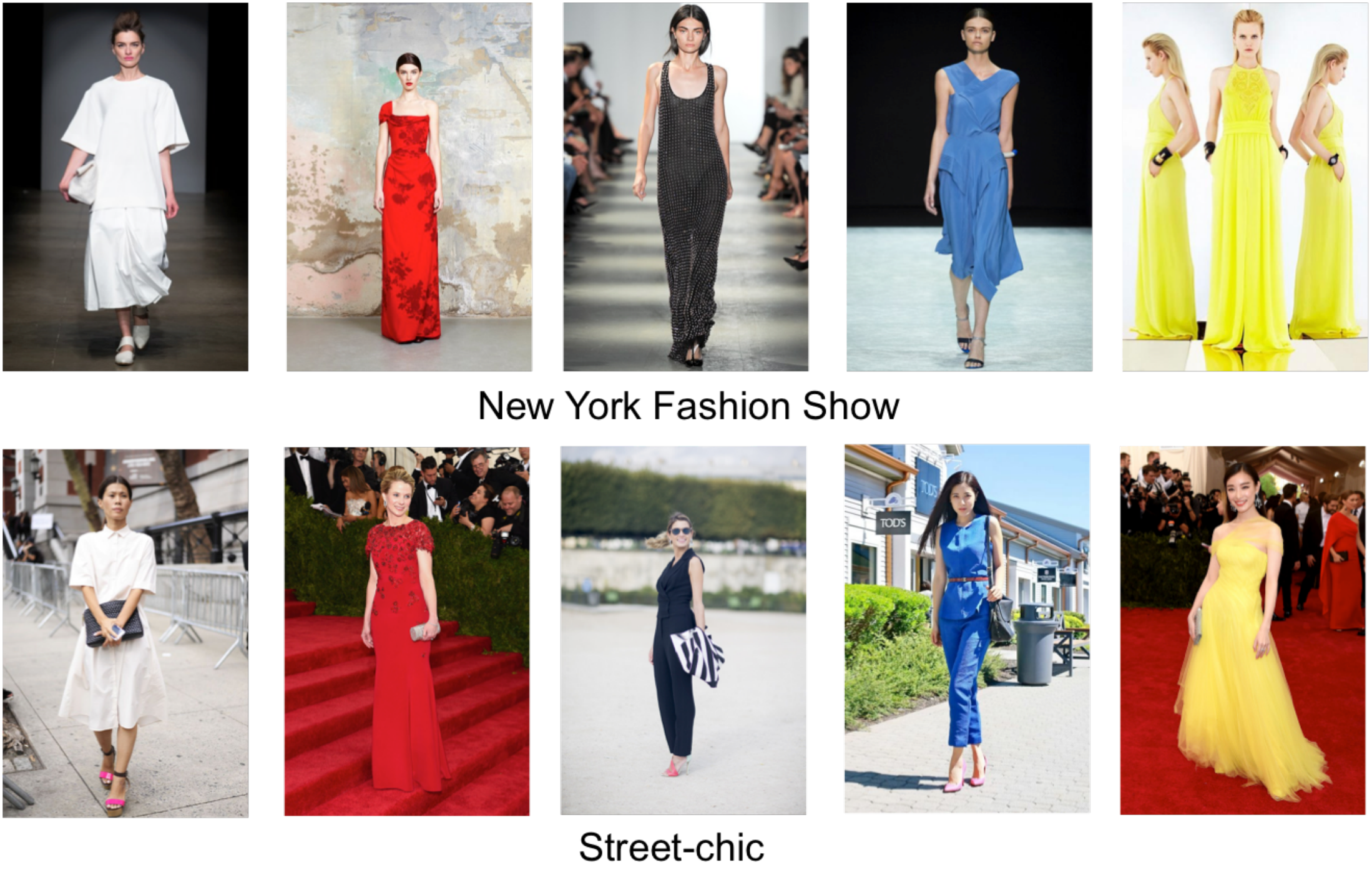, height=1.7in, width=3.3in}
\caption{The examples of the similarity of fashion trends shown both in (a) New York Fashion Show and (b) Street-chic.\label{fig:eval_example}}
\end{figure}

In order to analyze the influence from the New York Fashion Shows on people's daily life, we analyze the clothing attributes from the street chic images. As we mainly focus on the spring and summer fashion show influence, we collected 471 street chic images in 2014 and 575 street chic images in 2015 that were taken in New York from April to July and utilize these images as our street-chic data for fashion influence analysis. The changes of these attributes between 2014 and 2015 are shown in Figure \ref{fig:eval}. As shown in Figure \ref{fig:eval}, some changing trends of attributes are the same in fashion shows and in street chic. For example, there is a significant rise in both placket and tank top of the style category. In the pattern category, striped pattern in the upper body increased substantially. In the color category, blue in the lower body, blue, cyan, red and multicolor in the upper body showed an upward trend in both fashion show and street chic images. Consequently, we could tell that many people have the penchant to emulate clothing styles shown in the fashion shows. Some examples are shown in Figure \ref{fig:eval_example}. Interestingly, there are some different changes of fashion elements between the fashion shows and street chic images. For instance, we observe that the difference between the fashion shows and street chic lies in the long knee sleeve and outwear. We discovered that summer 2014 is the coldest in a decade, and people tended to wear outwear to keep warm. Another interesting observation is that people tend to wear bright-colored clothes in both the upper body and the lower body instead of wearing in only the upper body as shown in the fashion shows. A reasonable explanation is that light-colored clothes can absorb less sun light and keep people cooler; therefore, people are tempted to wear light color clothes in order to keep them cooler in the hotter weather in 2015. In summary, it seems clear that fashion trends from fashion shows indeed provide people a clothing reference and have significant influence on people's daily life. However, social condition and natural conditions, such as weather, could be important factors for people to determine what they are likely to wear. Furthermore, analysis of street chic images also has benefit to inspire fashion designers to come up with more main-stream designs.

\section{Conclusions}
In this work, we construct a large dataset from the New York Fashion Shows and street-chic in order to investigate the possible clothing fashion trends in New York. In addition, we investigate a learning based method to automatically detect fashion attributes as the representation of fashion trends. We further demonstrate that the proposed framework is effective in fashion trend discovery. Furthermore, we compare the analysis result from the New York Fashion Shows and street-chic images in order to investigate whether and to what extent the fashion shows have influence on New York people. In the future, we plan to collect more images to increase the data set for more comprehensive studies. Moreover, there is also keen interest in exploring the proper fashion outfits to recommend users by aggregating user preference in clothing style and one future direction would be to incorporate these elements into the existing model and the scope will be extending and applying our system to practical applications, for example clothing style and product recommendation. Besides, the concept of clustering will also be one of the options as it will be interesting to automatically learning unseen clothing elements \cite{hidayati:TrendNY}. 

\section{Acknowledgments}
This work was generously supported in part by Google, Yahoo,  Adobe, TCL, and New York State CoE CEIS and IDS. KuanTing Chen is supported by a scholarship by the Taiwan Government NSC Study Abroad Program grants 103-2917-I-002-159.

\bibliographystyle{abbrv}
\bibliography{sigproc}  
%
%

\end{document}